%% file: coling_latex.tex
\pdfoutput=1

\documentclass[11pt]{article}

\usepackage[final]{coling}

\usepackage{times}
\usepackage{latexsym}

\usepackage[T1]{fontenc}

\usepackage[utf8]{inputenc}

\usepackage{microtype}

\usepackage{inconsolata}

\usepackage{graphicx}

\usepackage{amsmath}
\usepackage{tabularx}
\usepackage{multirow}
\usepackage{url}
\usepackage{enumitem}
\usepackage[noabbrev]{cleveref}
\usepackage{booktabs}
\usepackage{tabularx}
\usepackage{caption}
\usepackage{array}
\usepackage{graphicx}
\usepackage{scalerel,xparse} 
\NewDocumentCommand\emojijap{}{\scalerel*{\includegraphics{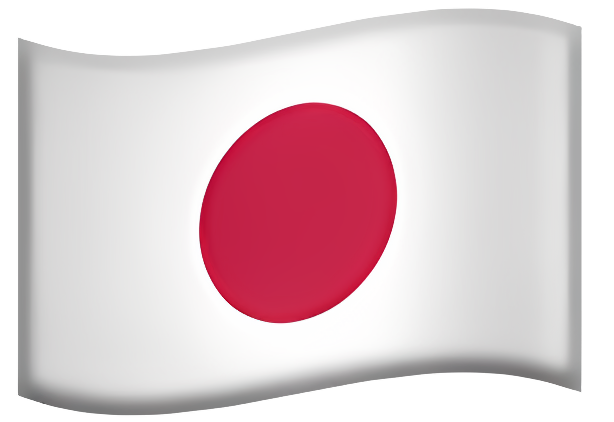}}{X}}
\NewDocumentCommand\emojipol{}{\scalerel*{\includegraphics{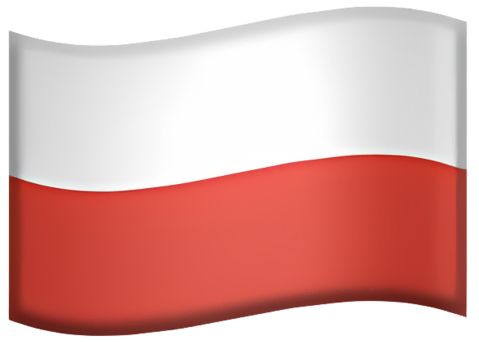}}{X}}
\NewDocumentCommand\emojius{}{\scalerel*{\includegraphics{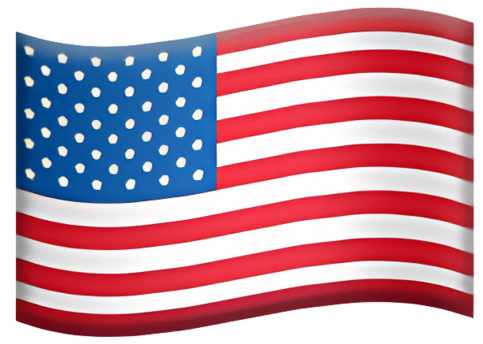}}{X}}
\usepackage{listings}
 \title{Context-Informed Machine Translation of Manga \\ using Multimodal Large Language Models}

\author{Philip Lippmann$^{\clubsuit *}$ \hspace{0.mm} Konrad Skublicki$^{\clubsuit}$ \hspace{0.mm} Joshua Tanner$^{\diamondsuit}$ \\ \hspace{0mm} \textbf{Shonosuke Ishiwatari}$^{\diamondsuit}$ \hspace{0mm} \textbf{Jie Yang}$^{ \clubsuit}$
\vspace{1mm} 
\\
$^{\clubsuit}$ Delft University of Technology \hspace{1mm}
$^{\diamondsuit}$ Mantra Inc. \hspace{1mm}\\
}

\begin{document}
\maketitle
\begin{abstract}
Due to the significant time and effort required for handcrafting translations, most manga never leave the domestic Japanese market.
Automatic manga translation is a promising potential solution. 
However, it is a budding and underdeveloped field and presents complexities even greater than those found in standard translation due to the need to effectively incorporate visual elements into the translation process to resolve ambiguities.
In this work, we investigate to what extent multimodal large language models (LLMs) can provide effective manga translation, thereby assisting manga authors and publishers in reaching wider audiences. 
Specifically, we propose a methodology that leverages the vision component of multimodal LLMs to improve translation quality and evaluate the impact of translation unit size, context length, and propose a token efficient approach for manga translation.
Moreover, we introduce a new evaluation dataset -- the first parallel Japanese-Polish manga translation dataset -- as part of a benchmark to be used in future research.
Finally, we contribute an open-source software suite, enabling others to benchmark LLMs for manga translation.
Our findings demonstrate that our proposed methods achieve state-of-the-art results for Japanese-English translation and set a new standard for Japanese-Polish.\footnote{Data and code available at: \url{https://github.com/plippmann/multimodal-manga-translation}. \\ *All correspondence: \texttt{p.lippmann@tudelft.nl}.}
\end{abstract}

\input{sections/intro}
\input{sections/related}
\input{sections/method}
\input{sections/experiments}

\input{sections/results}
\input{sections/conclusion}

\section*{Acknowledgments}
This work was supported by an Oracle
for Research Grant Award, as well as SURF Grant EINF-8535.

\bibliography{custom}

\appendix
\input{sections/app-proc}
\input{sections/app-cod}

\input{sections/app-mqm}
\input{sections/app-prompts}
\input{sections/app-ab}

\end{document}

%% file: sections/intro.tex
\section{Introduction}
A Japanese style of comics -- referred to as \emph{manga} -- has been popular with audiences outside of Japan for decades.
Handcrafting high quality translations, key to distributing manga world wide, is a difficult undertaking that takes significant time and effort.
As such, most manga never leave the domestic Japanese market.
Additionally, readers who do not speak a language into which manga is typically translated have limited or no access at all due to the high initial costs of translations.

\begin{figure*}[t]
    \centering
    \includegraphics[width=1\textwidth]{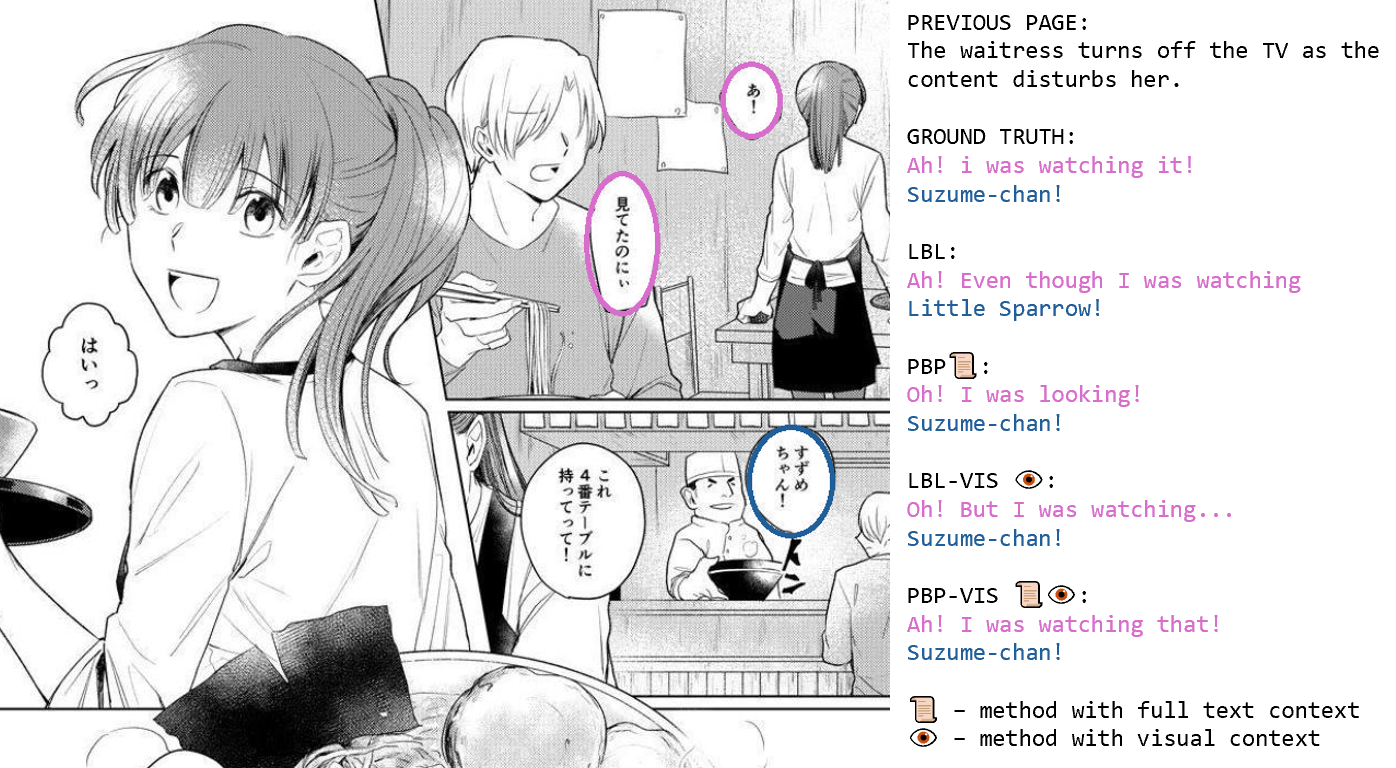}
    \caption{Comparison of translation outputs for methods with different context types. The preceding scene visually shows a TV, giving context to the complaints in purple. The previous and current scenes are set in a restaurant, making it improbable that ``Suzume'' refers to a sparrow rather than being a name.
    \copyright Kira Ito}
    \label{fig:modality_comparison}
\end{figure*}

The use of Neural Machine Translation (NMT) promises seamless translations from one language to another without involving a human translator~\citep{sutskever2014learning, aiayn}.
Still, successful applications of NMT to manga -- or comics in general -- remain limited, and automatic methods remain far from being able to reliably translate manga at a level comparable to humans~\citep{mantraold}.
This is in part due to the unique requirements of manga as a translation problem, which involves literary translation, handling split sentences across multiple speech bubbles, and especially resolving ambiguities using visual information.
For example, in \cref{fig:modality_comparison}, achieving an accurate translation requires integrating both textual and visual context from the current and preceding scenes.
%
%
%
%
%

Research into manga-specific NMT methods is limited, focusing mainly on Japanese-English translation due to a lack of parallel corpora for other language pairs~\cite{mantraold, mantranew}. 
Of these, only one method has attempted incorporating visual context into a model via a limited number of descriptive tags, yielding inconclusive results~\cite{mantraold}.
Previously proposed models were trained on a private JA-EN data set, which is not shareable due to copyright~\cite{mantraold, mantranew}.
Although there exist several general purpose manga data sets, such as Manga109~\cite{manga109_orig, manga109_2}, so far only one manga translation data set has been published for research purposes: OpenMantra~\cite{mantraold}.
However, its limited size makes it effectively an evaluation data set only, making it challenging to train models.

Large language models (LLMs) have shown to be capable translators across languages~\cite{lyu2023new, hendy2023good}. 
%
%
The release of multimodal LLMs -- those that make use of visual information in addition to text -- makes translation of media with visual nuance a possibility~\cite{lyu2024paradigm}.
This potentially bypasses the need for large parallel manga corpora for each language pair as LLMs do not need to be finetuned by the user.
Still, it is not clear how to best use these as effective manga translators. 

In this paper, we present a translation methodology using a multimodal multilingual LLM.
We evaluate a range of LLM-based translation approaches to empirically assess the impact of visual context, translation unit size, and context length.
We do this using an existing JA-EN manga data set and a new Japanese-Polish data set created for this purpose.
The JA-PL translation direction is chosen for its unique challenges, particularly due to the significant differences in syntactic structures and semantic nuances between Japanese, English, and Polish, and to address a low-resource language that nonetheless has a market for manga~\cite{swieczkowska2017towards}.
%
%
Finally, we contribute an open-source manga translation evaluation suite that allows users to choose the granularity of available context, provides automatic evaluation metrics, and enables testing of different LLMs.
%
%
%

In summary, our contributions are as follows:
\begin{itemize}[noitemsep,topsep=0pt]
    \item An LLM-based multimodal manga translation methodology that achieves state of the art results on JA-EN translations and can serve as a baseline for low-resource languages.
    \item An annotated set of 400 professionally translated manga pages (3705 sentences) that make up the first ever parallel JA-PL manga translation benchmark data set, as well as the largest manga translation data set to date.
    \item The first publicly available automatic manga translation evaluation software suite.
\end{itemize}

%% file: sections/related.tex
\section{Related Work}


\subsection{Automatic Methods for Manga}
Up to this point, the development of automatic manga translation methods that incorporate multimodal context has been limited.
\citet{mantraold} first proposed an NMT system for manga that makes use of contextual information obtained from images to inform the translation.
Their method is restricted to a single frame of context and the visual information obtained from the images is limited to 512 predefined labels.
Further work has explored the use of an additional frame or manga metadata to improve translation quality~\citep{mantranew}, however, without visual context.
Instead, we propose taking additional textual content of up to the entire manga volume into account to improve translations, as well as the full manga image without predefined labels.
Outside of translations,~\citet{chen2019multilingual} propose a sentiment analysis method on manga text and~\citet{guo2023m2c} propose an approach that makes use of both visual and textual modalities to complete empty speech bubbles in existing manga.
There has been sparse early-stage research into automatic methods for similar media, such as graphical novels~\citep{harshavardhan2024future} and American comics~\citep{hapsani2017optical}.

\subsection{Large Language Model Translations}
Translation using LLMs is appealing due to their ability to generate high-quality translations for various language pairs without the need for training on extensive parallel corpora or fine-tuning~\cite{he2023exploring}.
LLMs have previously been shown to be capable translators~\citep{wang2023document}, as well as evaluators of translation quality~\citep{kocmi2023large}.
Further, paragraph translations performed by LLMs have been shown to be effective when using basic English prompts at the sentence level~\citep{biao2023prompt}.
We propose multiple translation approaches and evaluate the quality of our LLM manga translations compared to finetuned transformer models and explore a low-resource language pair, JA-PL, as well as contribute a data set for evaluation.

\subsection{Multimodal Machine Translations}
Translating text embedded in images has been extensively explored in research~\citep{zhu-etal-2023-peit,lan2023exploringbettertextimage}.
Multimodal machine translation (MMT) has so far mainly been applied to translating image captions, outperforming the text-only baseline by leveraging additional visual information~\citep{gwinnup2023survey}.
MMT typically uses a single image with its corresponding text description as input~\citep{elliott2016multi30k}.
We investigate to what extend an increased visual context length is effective.
A further challenge comes from the discrepancy between the natural images and their description used to train vision encoders used for MMT and manga images, as manga has a unique hand-drawn art style with relevant text drawn into the image~\citep{guo2023bridging}.
Additionally, little attention has been paid to low-resource languages, with the vast majority of MMT research focused on the most popular language translation pairs~\citep{guo2022lvp, huang2023low}.
More recent LLMs have additional multimodal capabilities~\citep{huang2023kosmos, yin2023survey}, enabling them to perform MMT, though this has not been explored for the manga use case.
%

%% file: sections/method.tex
\section{Methodology}
\label{sec:method}

In this section, we first outline manga terminology, then present the problem, and finally introduce our LLM-based translation approaches that take advantage of multimodality and a longer context.

\begin{figure}[t]
    \centering
    \includegraphics[width=0.49\textwidth]{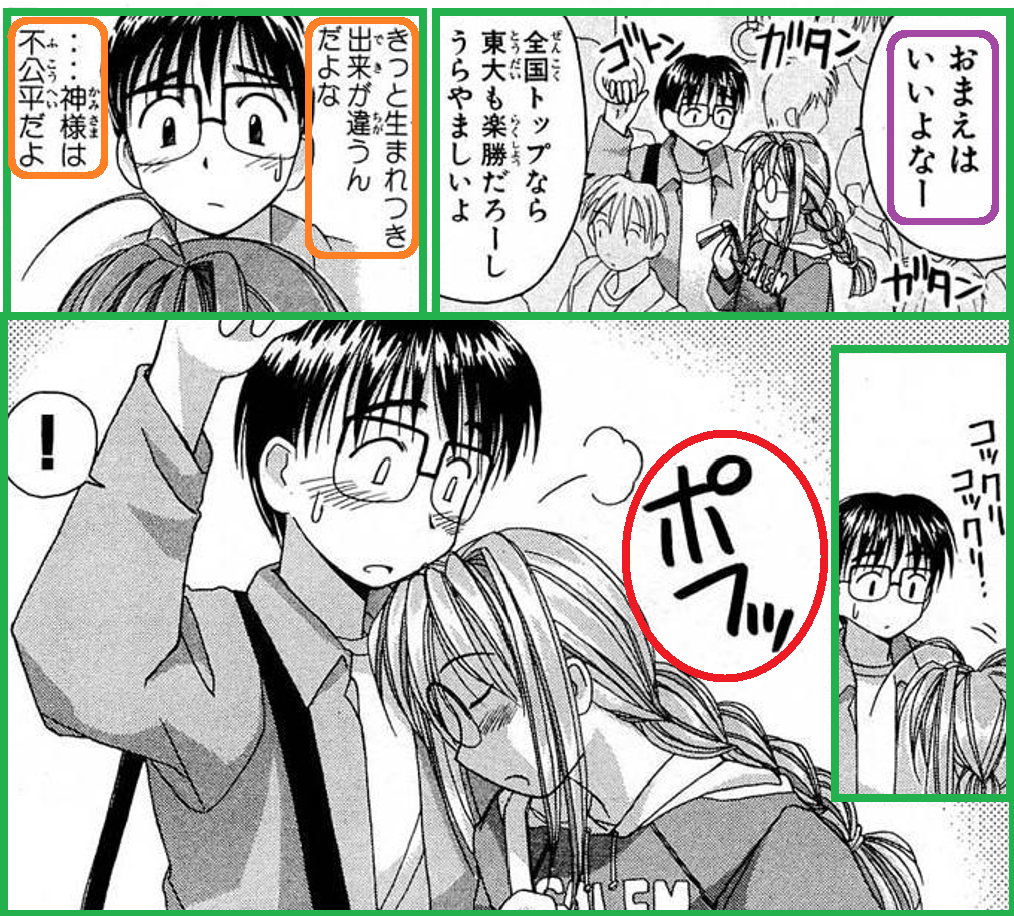}
    \caption{A manga page: 
    panel borders (\textbf{\textcolor{green}{green}}), example lines in speech bubbles (\textbf{\textcolor{purple}{purple}}), free flowing text (\textbf{\textcolor{orange}{orange}}) and sound effects (\textbf{\textcolor{red}{red}}).
    Courtesy of Akamatsu Ken, \copyright Kodansha}
    \label{fig:page_elements}
\end{figure}

\subsection{Terminology \& Problem Formulation}
Page-to-page manga translation involves three steps: (1) \emph{page processing} to identify elements on the page, detect text, and estimate reading order; (2) \emph{translating} the text into the target language; and (3) \emph{typesetting} the translated text onto the page in stylized font.
The focus of this paper is on (2), but we will discuss (1) and (3) in~\cref{sec:proc_typeset}.

A manga page consists of multiple story panels, referred to simply as \emph{panels}, as shown in Figure \ref{fig:page_elements}. 
Panels often contain text, which can be enclosed in a \emph{speech bubble} for text spoken or thought by characters, or free-flowing for background noise or sound effects. 
%
%
The term \emph{line} will always refer to the content of one speech bubble, narrative box, or cluster of free-flowing text.
%

For multimodal manga translation, we make use of the \emph{image} of the drawings on a single manga page, such as \cref{fig:page_elements}, which contains lines of text.
We make the assumption that the text contained on the page has already been recognized and is available.
Our goal is to obtain the correctly translated text for each line.
%
%

\subsection{Translation Approaches}
We use a variety of translation approaches -- summarized in \cref{tab:methods_overview} -- to assess the impact of multimodality, translation unit size, and context length and find the most performant configuration.
To establish a baseline, our first approach is a simple line-by-line approach (\texttt{LBL}).
This means that the model receives one single line to translate at a time, without any additional context about the manga it is translating.
Previous research has shown that LLMs perform better on translation tasks when given the entirety of a text compared to snippets, as they are able to incorporate the broader context more effectively~\cite{karpinska2023large}.
As such, the second approach we evaluate is page-by-page (\texttt{PBP}), where the model is given all lines from a given page in the correct reading order and outputs all the corresponding translations.

\begin{figure}[t]
    \centering
    \includegraphics[width=0.3\textwidth]{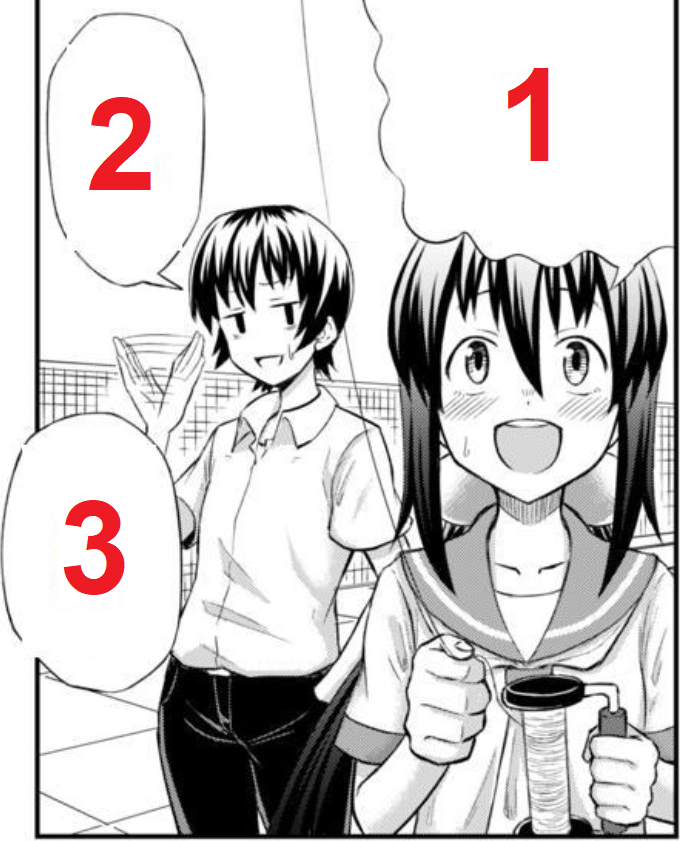}
    \caption{Fragment of a page annotated for the \texttt{PBP-VIS-NUM} method
    \copyright Mitsuki Kuchitaka.}
    \label{fig:page_annotated_for_pbp_num_vis}
\end{figure}

\subsubsection{Multimodal Translation}
Ideally, we would provide the LLM with just the image and it would recognize the text, obtain the visual context, and perform the translation. 
However, current models are not capable of this.
Instead, we investigate approaches where the model is given the lines on a page to translate, along with the corresponding image as additional visual context, enabling multimodal translation.

The first approach we investigate is the multimodal equivalent to \texttt{LBL}, referred to as \texttt{LBL-VIS}, where the model receives lines and the corresponding page image as visual context. 
The second approach utilizing visual context is \texttt{PBP-VIS}, which involves giving the model the entire text from one manga page and the page itself as an image.

The final approach aims to directly address the issues that multimodal LLMs have with reading non-Latin scripts.
The setup is the same as \texttt{PBP-VIS}, but the image of the manga page is modified to avoid the LLM performing any optical character recognition (OCR).
The contents of the speech bubbles in the image are removed and replaced with numbers indicating the reading order and corresponding to the list of Japanese lines the model is given to translate (see~\cref{fig:page_annotated_for_pbp_num_vis}). 
We call this approach \texttt{PBP-VIS-NUM} and it enables the model to locate the speech bubble more easily and relate its content to the exact panel in which it was placed, without performing any OCR on the text itself.

\begin{table}[t]
    \small 
    \centering
    \begin{tabularx}{\columnwidth}{>{\raggedright\arraybackslash}p{1.8cm} >{\centering\arraybackslash}p{1.5cm} >{\centering\arraybackslash}p{1.5cm} >{\centering\arraybackslash}p{1.5cm}}
    \toprule
    \textbf{Approach} & \textbf{Translation Unit} & \textbf{Textual Context} & \textbf{Visual Context} \\
    \midrule
    \texttt{LBL} & line & line & $\times$ \\
    \texttt{PBP} & page & page & $\times$\\
    \texttt{LBL-VIS} & line & line & page\\
    \texttt{PBP-VIS} & page & page & page\\
    \texttt{PBP-VIS-NUM} & page & page & num. page\\
    \texttt{VBP-VIS-COD} & page & page + sum. & page\\
    \texttt{VBP-VIS-3P} & page & 3 pages & 3 pages\\
    \texttt{VBP-VIS-ALL} & page & vol. + trans. & volume\\
    \texttt{VBV-VIS} & volume & volume & volume\\
    \bottomrule
    \end{tabularx}
    \caption{Overview of the proposed approaches. Abbreviations: ``num.'' is numbered, ``sum.'' is summary, ``vol.'' is volume, and ``trans.'' is translation so far.}
    \label{tab:methods_overview}
\end{table}

\subsubsection{Long-Context Translation}
%
Intuitively we want to make use of context lengths exceeding single lines or pages to adequately capture evolving story lines and character development and accurately translate entire stories in an internally consistent way.
The remaining approaches we present are designed to address this.

The first of these multi-page approaches provides the model with the previous and next page as additional context to give more information to the LLM. 
We refer to this as \texttt{VBP-VIS-3P}, as it translates the volume sequentially one page at a time (\texttt{VBP}), using visual context (\texttt{VIS}), and using three pages' worth of context (\texttt{3P}). 
Going a step further, we explore \texttt{VBP-VIS-ALL}, where the model is provided with the images and lines from an entire manga volume, as well as all the translations so far, and queried to translate the next untranslated page.
This process is repeated sequentially for every page in the volume.

As the input and output length increases, the limited context windows of LLMs are quickly exhausted and performance is diminished~\citep{liu2023lost}.
To overcome this, we introduce the scalable \texttt{VBP-VIS-COD} approach, where we extend chain of density summarization (COD)~\cite{adams2023sparse} to keep a rolling, fixed-length summary of the story's developments as our context.
Besides the image and its corresponding text, the model is given a summary of the story thus far in the target language as additional context.
For a detailed overview of this approach, see~\cref{sec:cod}.
The last evaluated approach, \texttt{VBV-VIS} translates an entire manga volume in a single call.
Similar to \texttt{VBP-VIS-ALL}, we provide the LLM with the texts and images from an entire manga volume, but then instruct it to respond with the translations for the entire volume in a single query.

\begin{table}[t]
    \centering
    \small
    \newcolumntype{C}{>{\centering\arraybackslash}X}
    \begin{tabularx}{\columnwidth}{l C C C}
    \toprule
    \textbf{Manga Title} & \textbf{Genre} & \textbf{\# Pages} & \textbf{\# Lines}\\ 
    \midrule
    \emph{Balloon Dream} & Romance & 38 & 314 \\
    \emph{Boureisougi} & Mystery & 36 & 274\\
    \emph{Rasetugari} & Fantasy & 54 & 359\\ 
    \emph{Tencho Isoro} & SoL & 40 & 311 \\ 
    \emph{Tojime no Siora} & Battle & 46 & 334\\ 
    \bottomrule
    \end{tabularx}
    \caption{Overview of the OpenMantra data set \cite{mantraold}. SoL is slice-of-life.}
    \label{tab:openmantra_overview}
\end{table}

%% file: sections/experiments.tex
\section{Experiments}
\label{sec:exp}

\subsection{Data}
Machine translation of manga is a niche field with little publicly available data and no established research benchmarks~\cite{mantraold}.
There exists a plethora of manga data that could be used for training and evaluating machine translation systems~\cite{swieczkowska2017towards, sachdeva2024manga}. 
However, from a research perspective, the main issue with manga is that, due to its commercial nature, most manga is protected by Japanese and local copyright laws~\cite{schroff2019alternative}.
Previous manga-related works have addressed this problem in different ways. 
Some researchers resort to using private data sets~\cite{rigaud2021text, mantraold, mantranew}, while others use the very few publicly available copyright-free manga, accepting the trade-off of unlabeled data~\cite{sharif2021effective}.

To date, there has been only one manga translation data set made public for research purposes -- the OpenMantra data set by~\citet{mantraold}.
It consists of five independent Japanese-language manga volumes, totaling 214 pages (1593 speech bubbles).
Details of this data set are shown in~\cref{tab:openmantra_overview}.
Each volume in this data set has annotations for the locations of panels and text boxes on the page, as well as the contents of the text boxes, and the reading order, with professional translations into English and Chinese. 
We use this data set to evaluate JA-EN translations, splitting it into two parts: validation set (\emph{Balloon Dream} and \emph{Tojime no Siora}) and test set (\emph{Boureisougi}, \emph{Rasetugari}, and \emph{Tencho Isoro}).
%

\subsection{New Japanese-Polish Manga Data Set}
In addition to JA-EN, we explore JA-PL translation; as English and Polish belong to different language families, diverge significantly in terms of morphology, and have different grammatical structures.
We provide professional JA-PL translations of the slice-of-life manga \emph{Love Hina} to create a data set for research purposes.
We make volumes 1 and 14 available and our annotation process closely follows the existing annotations of the Japanese text. 
The newly contributed data set contains 400 pages and 3705 individual lines (\emph{i.e.} speech bubbles, sound effects, etc.) split across the two volumes and is distributed as a set of images, corresponding to one image per page, and the corresponding metadata containing original and translated text, as well as their coordinates on the page.
This exceeds the previously largest manga translation data set, OpenMantra~\cite{mantraold}, in size.
We propose a 50:50 validation:test split for this data set, using the first volume (200 pages and 1810 lines) as the test set and the last volume (200 pages and 1895 lines) as the validation set.
This decision is motivated primarily by the fact that the first volume establishes the story, providing a fairer benchmark for the long-context methods, as opposed to the last volume, which depends on unavailable context, being the 14th installment in the series.

Our annotation process closely follows the existing annotations of the Japanese text.
The original lines were matched with the corresponding translated lines primarily based on location, and if impossible, based on content. 
However, in edge cases the Polish edition left very small text untranslated as a stylistic choice.
%
The reading order was first estimated using the tool provided by~\citet{sachdeva2024manga} and then corrected by hand based on the actual speech bubbles.
During the annotation process, we noticed several characteristics of this title and the unique challenges it presents for translation.
Some characters in \emph{Love Hina} speak the Kansai dialect of Japanese. 
According to the literature, there is no consensus on how to translate this dialect into Polish, with different translators choosing different Polish dialects~\cite{jaskiewicz2020reprezentacja}. 
Another challenge is that one of the secondary characters speaks in a manner resembling samurai speech -- a common trope in manga~\cite{duc2019znaczenie}. 
Again, there is no consensus on how to convey this in Polish.
As such , users of the data set should be aware that some ``incorrect'' translations may be just as valid in these cases.

\subsection{Baselines}
\label{sec:eval_base}
We employ four baseline methods for JA-EN translations. 
The first two baselines, \texttt{Scene-NMT} and \texttt{Scene-NMT-VIS}, come from the original automatic manga translation work by~\citet{mantraold}. 
The first method uses a transformer-based model to translate the contents of entire panels at once without multimodal context, while the second method includes visual features as well.
The third baseline method we use -- and current state-of-the-art for automatic manga translation -- is \texttt{Scene-aware-NMT}~\cite{mantranew}, which translates manga panel by panel as well, using a transformer-based model but uses the text from the previous panel as additional context.
The translation outputs for all these previously listed methods were kindly provided to us by the authors of the respective works.
This allowed us to use our own data splits and ensure that all methods were evaluated equally and comparably.

The last baseline method we use is Google Translate (\texttt{GT}) due to its support for a wide range of languages and availability.
\texttt{GT} is our only baseline for JA-PL translations.
All \texttt{GT} translations were carried out in April and May 2024, using the Google Translate API with the corresponding Python library.\footnote{\url{https://pypi.org/project/googletrans/}}

\begin{table*}[t]
   \centering
   \begin{tabularx}{\textwidth}{@{\extracolsep{\fill}}lcccccccccc@{\extracolsep{\fill}}}
   \toprule
   \multirow{2}{*}{\textbf{Method}} & \multicolumn{4}{c}{\emojijap \ \textbf{JA-EN} \emojius} & & \multicolumn{4}{c}{\emojijap \ \textbf{JA-PL} \emojipol} \\
   \cmidrule(lr){2-5} \cmidrule(lr){7-10}
    & \textbf{ChrF} & \textbf{BRTS} & \textbf{BLRT} & \textbf{xCMT} & & \textbf{ChrF} & \textbf{BRTS} & \textbf{BLRT} & \textbf{xCMT} \\ 
   \midrule
   \texttt{GT}  & 34.2 & 0.895 & 0.525 & 0.729 & & 22.3 & 0.826 & 0.446 & 0.457 \\
   \texttt{Scene-NMT}  & 34.2 & 0.897 & 0.512 & 0.651 & & - & - & - & - \\
   \texttt{Scene-NMT-VIS}  & 34.5 & 0.895 & 0.507 & 0.664 & & - & - & - & - \\
   \texttt{Scene-aware-NMT}  & 36.1 & \textbf{0.903} & 0.534 & 0.670 & & - & - & - & - \\
   \midrule
   \texttt{LBL}  & 32.7 & 0.883 & 0.523 & 0.716 & & 24.2 & 0.844 & 0.495 & 0.531 \\ 
   \texttt{PBP}  & 36.0 & 0.898 & 0.565 & 0.758 & & 25.6 & \textbf{0.852} & 0.538 & 0.565 \\ 
   \midrule
   \texttt{LBL-VIS}  & 35.6 & 0.900 & 0.551 & 0.746 & & 24.9 & 0.845 & 0.515 & 0.543 \\ 
   \texttt{PBP-VIS} & 36.6 & \textbf{0.903} & 0.581 & \textbf{0.776} & & 25.6 & \textbf{0.852} & \textbf{0.539} & \textbf{0.567} \\ 
   \texttt{PBP-VIS-NUM}  & \textbf{36.8} & 0.900 & \textbf{0.582} & \textbf{0.776} & & \textbf{25.7} & 0.851 & 0.532 & 0.566 \\
   \midrule
   \texttt{VBP-VIS-COD}  & 35.9 & 0.900 & 0.566 & 0.769 & & 25.1 & 0.846 & 0.523 & 0.550 \\
   \texttt{VBP-VIS-3P}  & 35.9 & 0.897 & 0.565 & 0.754 & & 25.6 & 0.843 & 0.530 & 0.559 \\
   \texttt{VBP-VIS-ALL}  & 35.7 & 0.893 & 0.556 & 0.760 & & 24.9 & 0.840 & 0.521 & 0.561 \\
   \texttt{VBV-VIS}  & 34.9 & 0.884 & 0.539 & 0.733 & & 24.5 & 0.833 & 0.510 & 0.534 \\ 
   \bottomrule
   \end{tabularx}
   \caption{Performance metrics for all approaches for JA-EN and JA-PL translation. Best scores for each translation direction are in \textbf{bold}. BRTS refers to BERTScore, BLRT to BLEURT, and xCMT to xCOMET.}
   \label{tab:results}
\end{table*}

\subsection{Automatic Evaluation}
\label{sec:eval_metrics}
For evaluation, we use a range of automated metrics applied at the sentence level. 
We use a lexical n-gram matching heuristic metric in ChrF~\cite{popovic2015chrf}.
%
Although the reliability of this type of metric has been questioned over the years~\citep{thai2022exploring}, they remain among the most widely used in machine translation~\cite{mathur2020tangled, kocmi2024navigating}. 
%
%
ChrF provides scores on a scale from 0 to 100, where higher scores indicate higher quality translations.

The first non-lexical machine translation evaluation metric we use is BERTScore~\cite{zhang2019bertscore}, considered a good representative of the embedding-based metrics category~\cite{saadany2021bleu}. 
Although not perfect, it has been shown to detect important content words and is well suited to score candidates from different languages~\cite{hanna2021fine}. 
%
%
Next, we report scores with a learned metric, BLEURT~\cite{sellam2020bleurt}, specifically the top-performing BLEURT-20 model~\cite{pu2021learning}. 
%
%
%
The last metric we report is the learned metric xCOMET~\cite{guerreiro2023xcomet}, specifically xCOMET-XXL. 
xCOMET is an open-source learned metric that performs error span detection in addition to standard sentence-level evaluation.
It is currently considered the best-performing publicly available metric~\cite{freitag2023results}. 
Among all the metrics we employ, it is the only one that calculates its score based not only on the references and hypotheses but also on the source text. 
%
%
BERTScore, BLEURT, and xCOMET return a score on a scale of 0 to 1, with results closer to 1 being preferable.

%

\subsection{Human Evaluation}
In addition to our extensive automatic evaluation, we perform a human evaluation with a professional JA-EN manga translator using the Multidimensional Quality Metrics (MQM) translation evaluation framework~\cite{burchardt2013multidimensional, ISO5060}.
We use MQM with a manga-specific list of issue types that cover different types of errors, such as accuracy, fluency, and style. 
A complete overview of our MQM process is shown in \cref{sec:mqm}.
Each error type is assigned a severity level, ranging from minor to critical, depending on the impact of the issue on overall quality. 
MQM provides a scoring system that allows for the calculation of overall quality scores based on the number of identified issues and their severity levels. 
These scores have an upper bound of 1 and no lower bound, with a higher score being preferable.
We choose the \emph{Tencho Isoro} manga for our MQM evaluation. 
We compare the official commercial translation of the manga, the \texttt{GT} baseline, and our best performing approach (\texttt{PBP-VIS}) to evaluate how a professional human translator would judge each.
%

\subsection{Prompting}
\label{sec:exp_prompts}
We follow the approach of previous works~\cite{hendy2023good, karpinska2023large, lyu2024paradigm} and investigate the out-of-the-box translation performance of GPT-4 Turbo \cite{openai2024gpt4}.
The specific version we use is \texttt{gpt-4-turbo-2024-04-09} at default hyperparameters with a temperature $T = 0.5$, accessed through the OpenAI Python library.\footnote{\url{https://github.com/openai/openai-python}}
For all multimodal translations, we append the relevant image(s) of the page(s) as a \emph{jpeg} file to the LLM query via its respective API.
We run each configuration once due to the high costs involved in sending entire manga volumes to commercial multimodal LLMs.
The complete prompts we use for every translation are shown in~\cref{sec:prompts}.
Each approach described in~\cref{sec:method} is evaluated one-shot, i.e., with one given example in the prompt.
We did not find a measurable difference between one-shot and five-shot prompting when evaluating on the validation data.
The model is always prompted in English -- regardless of the target language -- as this has yields the best results for LLM translations~\cite{zhang2023prompting}.
Based on experiments on the validation data, we ask the model to explain how the image influences the translation, ensuring that the visual context is taken into account.
%

\subsection{Manga Translation Evaluation Suite} 
We release our evaluation suite to advance research in automatic manga translation. 
It enables benchmarking of various LLMs by adjusting textual context, visual context, and translation unit size. 
The suite integrates all methods from \cref{sec:method} for comprehensive evaluation and facilitates automatic assessment using the four previously outlined metrics. 
With plug-and-play functionality, researchers can easily utilize existing data sets, including OpenMantra and ours, while introducing new prompts and exploring alternative LLMs.

%% file: sections/results.tex
\section{Results \& Discussion}

\subsection{JA-EN Translation}

We present our findings in~\cref{tab:results}.
Among the methods proposed in previous studies, \texttt{Scene-aware-NMT} demonstrates competitive performance, especially on BERTScore, surpassing other previous manga-focused translation methods, consistent with their reported findings.
However, our proposed methods show improvements across multiple metrics.
Our basic approach, \texttt{LBL}, performs slightly worse than \texttt{GT} in most aspects. 
The \texttt{PBP} method shows substantial improvement over \texttt{LBL}, outperforming all baselines on BLEURT (0.565) and xCOMET (0.758), confirming the potential of LLMs as manga translators, even without visual context.

\noindent\textbf{Visual Context}
The addition of visual context significantly improves scores across all metrics for both \texttt{LBL} and \texttt{PBP} methods. 
\texttt{PBP-VIS} and \texttt{PBP-VIS-NUM} achieve the best scores across most metrics, with \texttt{PBP-VIS-NUM} slightly outperforming on ChrF (36.8) and BLEURT (0.582). 
These results confirm that additional visual context significantly improves LLM translation quality, representing a novel approach in automatic manga translation.
Additionally, we perform an ablation study to clarify the role of key visual features, the results of which are discussed in~\cref{sec:ab}.

The results of our human evaluation are presented in \cref{tab:humaneval}. 
\texttt{PBP-VIS}, clearly outperforms the \texttt{GT} baseline in overall score. 
However, according to the MQM evaluation conducted by a single professional translator, \texttt{PBP-VIS} is more prone to errors than the official human manga translation.
While our method has fewer ``minor'' and ``major'' errors compared to the official translation, it exhibits a significantly higher number of ``critical'' errors.
These findings indicate that although our method establishes the current state of the art for automatic manga translation, human translation remains superior in quality. 
Although these metrics provide context for assessing our method's efficacy, translation quality is inherently subjective and challenging to measure. 
While our best translation scores lower than the official translation, we find it enjoyable to read and coherent -- a standard the \texttt{GT} translation does not meet.

\noindent\textbf{Context Length}
Interestingly, providing context beyond the page level does not enhance translation quality. 
\texttt{VBP-VIS-COD}, using only a short summary of previous events, performs better than other long-context methods across most metrics. 
Conversely, \texttt{VBV-VIS}, which translates the entire volume in one query, shows the lowest performance among our visual context methods. 
These findings suggest an inverse relationship between translation quality and input length beyond a single page for multimodal LLM translation. 
This counter intuitive result highlights the importance of optimizing input size for LLM-based translations.

\subsection{JA-PL Translation}
For the JA-PL data, we do not report the results of methods proposed by other authors, as these are not trained on Polish data and therefore perform poorly.
For JA-PL translations, we observe that across methods, scores are generally lower compared to JA-EN translations.
However, all our approaches significantly outperform the \texttt{GT} baseline.
Further, we note that our top performing methods, \texttt{PBP-VIS} and \texttt{PBP-VIS-NUM}, perform similarly on JA-PL to JA-EN.

\noindent\textbf{Visual Context and Context Length}
Again visual context improves performance, though to a much lesser extend than for JA-EN translation. 
For \texttt{PBP}, the impact of visual context is minimal, with \texttt{PBP-VIS} and \texttt{PBP-VIS-NUM} performing similarly to \texttt{PBP}.
Long-context approaches show mixed results, again performing worse than \texttt{PBP-VIS} and \texttt{PBP-VIS-NUM}.

\subsection{Implications and Broader Impact}
\texttt{PBP-VIS} and \texttt{PBP-VIS-NUM} consistently achieve the best results for both JA-EN and JA-PL translations.
The effectiveness of our methods across translations suggests broad applicability to different language pairs. 
The cross-lingual success of our methods indicates that the benefits of incorporating visual context in manga translation are language-independent.
Moreover, our \texttt{PBP-VIS} and \texttt{PBP-VIS-NUM} methods achieve the highest scores across all metrics, setting the state of the art for automatic manga translation.

Notably, we observe that translation quality does not necessarily improve with longer context, challenging common assumptions in machine translation. 
This finding aligns with previous research, which indicates that the quality of output from LLMs tends to diminish as the length of the input increases~\citep{levy2024tasktokensimpactinput}. 
The is contrary to the results we observe when additional visual context is taken into account.
To optimize performance when using LLMs for multimodal translations, it is advisable to prioritize smaller input sizes of a single page. 
Translation quality tends to deteriorate more significantly as the LLM processes longer text, even if it contains more information relevant to the story.

\begin{table}[t]
\centering
\begin{tabular}{@{}l@{\hskip 0.05in}r@{\hskip 0.15in}r@{\hskip 0.15in}r@{\hskip 0.15in}r@{}}
\toprule
\textbf{JA-EN} & \multicolumn{1}{c}{\textbf{Minor}} & \multicolumn{1}{c}{\textbf{Major}} & \multicolumn{1}{c}{\textbf{Critical}} & \multicolumn{1}{c}{\textbf{Score}} \\
\midrule
Official & \multicolumn{1}{c}{14} & \multicolumn{1}{c}{50} & \multicolumn{1}{c}{107} & \multicolumn{1}{c}{-1.31} \\
\texttt{GT} & \multicolumn{1}{c}{5} & \multicolumn{1}{c}{20} & \multicolumn{1}{c}{272} & \multicolumn{1}{c}{-4.25} \\
\texttt{PBP-VIS} & \multicolumn{1}{c}{8} & \multicolumn{1}{c}{18} & \multicolumn{1}{c}{160} & \multicolumn{1}{c}{-1.98} \\
\bottomrule
\end{tabular}
\caption{Human evaluation MQM results for JA-EN. Errors are number per category (lower preferable).}
\label{tab:humaneval}
\end{table}

%% file: sections/conclusion.tex
\section{Conclusion}
Our investigation of multimodal LLMs for automatic manga translation reveals significant advancements in this emerging field.
We evaluate various LLM-based translation approaches, considering text-only, image-informed, and volume-level contexts.
Leveraging the vision component of multimodal LLMs, we enhance translation quality by incorporating visual elements to resolve ambiguities.
However, we find that additional textual context does not consistently improve performance.
Our methodology achieves state-of-the-art results for JA-EN translations and sets a new standard for JA-PL translations.
We also introduce the first parallel JA-PL manga translation data set and an open-source benchmarking suite for LLMs.

\section{Limitations}
The first limitation of this study is the amount of data used for testing. 
While we make meaningful contributions to addressing this issue, there is still a severe lack of evaluation data, making it difficult to determine how consistent our findings would be across different authors and genres. 
Additionally, we only investigate one language other than English, constrained by our ability to manually inspect outputs and analyze model mistakes in other languages.

Related to this is the fact that some manga series span multiple volumes. 
Translations of later volumes in a series would undoubtedly benefit from including earlier volumes in the available context.
Due to the lack of suitable data, we limit ourselves to translations of single volumes, leaving multi-volume narratives to future work.

%
%
%

Finally, there are obvious limitations when using a commercial, closed-source LLM as we do in this paper, such as potential data leakage issues and the unlikely scenario that some of the manga used might have been part of the training data. 
Still, the availability and quality of open-source multimodal multilingual LLMs is very limited at this time, and as such we leave a study using alternatives to future work.

%% file: sections/app-proc.tex
\section{Page Processing and Typesetting}
\label{sec:proc_typeset}

\begin{figure*}[t!]
    \centering
    \includegraphics[width=\textwidth]{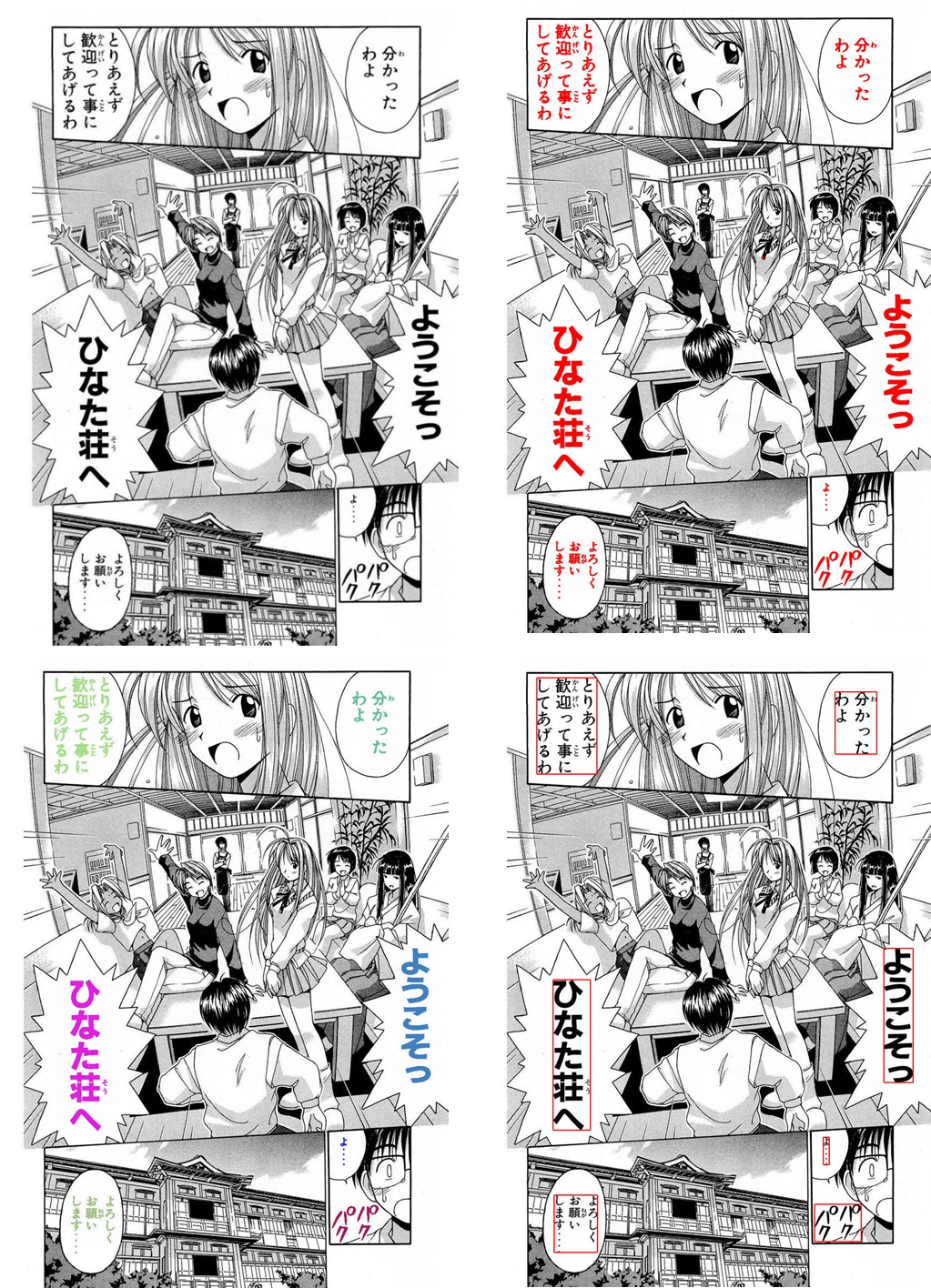}
    \caption{Stages of the text detection pipeline. First, pixels belonging to letters are identified. Then, the pixels are clustered into utterances. Lastly, bounding boxes are computed. Courtesy of Akamatsu Ken, \copyright Kodansha, from the Manga109-s dataset \cite{manga109_orig, manga109_2, manga109_building}}
    \label{fig:text_detection_pipeline}
\end{figure*}

\begin{figure*}[t]
    \centering
    \includegraphics[width=\textwidth]{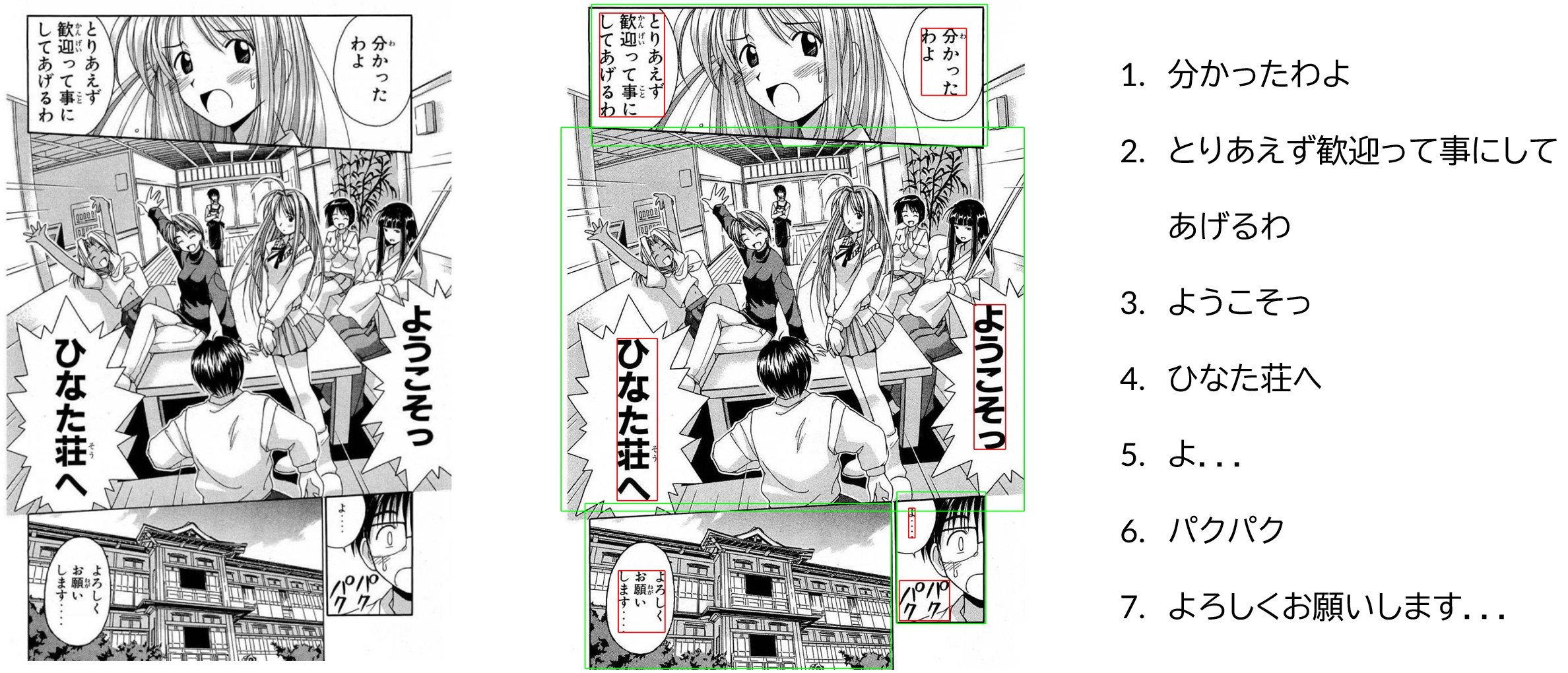}
    \caption{Page processing pipeline. The reading order is estimated based on the relative location of the detected panels (\textbf{\textcolor{green}{green}}) and text boxes (\textbf{\textcolor{red}{red}}). Courtesy of Akamatsu Ken, \copyright Kodansha, from the Manga109-s dataset \cite{manga109_orig, manga109_2, manga109_building}}
    \label{fig:page_processing_pipeline}
\end{figure*}

Page-to-page manga translation involves three distinct steps: (1) \emph{page processing} to identify elements on the page, detect text, and estimate reading order; (2) \emph{translating} the text into the target language; and (3) \emph{typesetting} by removing the source text from the page and inserting the translated text in a stylized font.
We will discuss (1) and (3) in this section.

\noindent \textbf{Page Processing}
The first step in manga translation is identifying the elements on the page. 
Here, we will present an example of a manga page processing pipeline composed of methods proposed by previous research and publicly available manga tools.
For text detection, we employ the unconstrained method proposed by~\citet{del2020unconstrained} to account for text that is not contained within speech bubbles -- see the top right of \cref{fig:text_detection_pipeline}.
However, before applying Optical Character Recognition (OCR) to the detected text fields, we need to group it into clusters belonging to the same utterance. 
To accomplish this, we apply a method inspired by~\citet{rigaud2021text} -- we utilize the OPTICS algorithm~\cite{ankerst1999optics}, specifically the Python pyclustering library implementation~\cite{Novikov2019}, to cluster the text -- see the bottom left of \cref{fig:text_detection_pipeline}. 
We then compute the bounding boxes of these obtained text clusters and discard those that are too small to contain text -- see the bottom right of \cref{fig:text_detection_pipeline}. 
Finally, we apply Manga OCR\footnote{https://github.com/kha-white/manga-ocr} for text recognition -- see \cref{fig:page_processing_pipeline}.

For panel detection and estimating the reading order, we utilize the Magi system~\cite{sachdeva2024manga}. 
In theory, Magi is capable of creating a transcript of a manga page independently, but it was trained on translations of manga and is not well-suited for Japanese text detection. 
As such, we only utilize some of its functionalities. 
A visualization of the page processing pipeline can be seen in Figure \ref{fig:page_processing_pipeline}. 
First, we use the process described previously to detect text boxes. 
Then, we employ Magi to detect panels and estimate the reading order based on the relative locations of text boxes and panels. 
Lastly, we utilize Manga OCR for text extraction.

\noindent \textbf{Typesetting}
The final step of a page-to-page manga translation -- inserting the translated text back into the image -- involves two steps: cleaning and lettering.
We did not perform it as part of our study but, for the sake of giving potential future works a comprehensive guide to follow, we will still outline the procedure here.
Cleaning refers to the removal step of the text, in which the original Japanese text is removed from the image used in the translation process. 
One could utilize an image inpainting model for this task~\cite{Nazeri_2019_ICCV}; the regions containing text lines are replaced by the inpainting model, which effectively removes the text even when it is overlaid on textured images or drawings. 
Alternatively, a method that performs this step specifically for manga has been proposed~\cite{mangajuicer}.

Lettering is the final step, where the translated text is rendered onto the cleaned image, with an optimized font size and placement that fits the manga aesthetic.
The location of the rendered text is chosen to maximize the font size while ensuring that all text remains within the designated text region. 
This step ensures that the translated text is legible and properly integrated into the image.
To the best of our knowledge, there exists no prior work that proposes to do this automatically at this time.
Though there are plenty of manual tools (both free and commercial) that make it possible to come up with a semi-automated approach if the coordinates of the original text are known.

%% file: sections/app-cod.tex
\section{Chain of Density Summarization}
\label{sec:cod}
\begin{figure*}[h!]
\begin{lstlisting}
Existing Summary from the previous Translation: {self.prev_context}

The most recent Observation from the {self.lang} translation was: {self.observation}

You will generate new increasingly concise, entity-dense summaries based on the above Existing Summary and most recent Observation.

Keep the summaries in {self.lang}.

You will create 3 summaries. You will create each of them by following the following two steps:
    -Step 1. If possible, identify 1-3 Informative Entities (";" delimited) from the most recent Observation which are missing from the Existing Summary.
    -Step 2. Write a new, denser summary of identical length which covers every entity, action, and detail from the previous Existing Summary plus the Informative Entities from the Observation.

An Informative Entity is:
    -Relevant: to the translation's unfolding narrative.
    -Specific: descriptive yet concise (10 words or fewer).
    -Novel: not in the previous summary.
    -Faithful: an accurate, detailed reflection of the translation.

Guidelines:
    -The first of the three summaries must be long (but less than ~{lmax} words) yet highly non-specific, containing little information beyond the entities marked as missing. Use verbose language and fillers (e.g., "In this part of the translation, the main character encounters ...") to reach ~{lmax} words.
    -Make every word count: rewrite the previous summary to improve flow and make space for additional Informative Entities.
    -Make every word count: rewrite the previous summary to improve flow and make space for additional Informative Entities.
    -Make space with fusion, compression, and removal of uninformative phrases like "the scenario presents".
    -The summaries should become highly dense and concise yet self-contained, e.g., easily understood without referencing the fact that a translation is being performed, and contain all information of the narrative thus far.
    -Informative Entities can appear anywhere in the new summary.
    -Only drop the least relevant Informative Entities from the previous summary if the summary length exceeds ~{lmax} words. Otherwise carry all previous Informative Entities to the new summary.

Answer in JSON. The JSON should be a list (length 3) of dictionaries under the key "summaries". Each dictionary should contain keys "Informative_Entities" (storing the Informative Entities included in the corresponding summary) and "Denser_Summary" (containing the summary).
\end{lstlisting}
\caption{Prompt used for Chain of Density summarization.}
\label{fig:promptCOD}
\end{figure*}

\noindent In the context of text summarization, Chain of Density (COD)~\cite{adams2023sparse} prompting can be used to generate high-quality summaries by breaking down the summarization task into a series of smaller, more manageable sub-tasks.
COD prompting can be applied to summarization as follows:

\noindent \textbf{Initial Prompt:} The process starts with a prompt that instructs the LLM to read and understand the source text that needs to be summarized.

\noindent \textbf{First Generation:} The LLM generates a brief summary or highlights the key points from the source text based on the initial prompt.

\noindent \textbf{Iterative Prompting:} The generated summary from the previous step is used as the prompt for the next step.
The LLM is then prompted to expand or refine the summary by adding more details, rephrasing certain parts, or reorganizing the information.
This step can be repeated multiple times, with each subsequent prompt building upon the previous summary.

\noindent \textbf{Final Summary:} After several iterations, the final summary should be a coherent, concise, and informative representation of the source text.

By breaking down the summarization process into smaller steps, COD helps the LLM maintain focus and context throughout the summary generation process.
This can lead to more coherent and accurate summaries, as the model can incrementally refine and improve the summary at each step.
It's important to note that the effectiveness of COD prompting for summarization may depend on the quality of the initial prompt, the complexity of the source text, and the LLM's capabilities.

The \texttt{VBP-VIS-COD} approach we propose does not try to make use of the large context window size of GPT-4 Turbo and instead uses COD to maintain a rolling summary of the developments in the story so far. 
In addition to the image, the model is also given a summary of the story so far in the target language as additional context. 
It is then asked to, in addition to the translation, return the description of the events taking place on the page being translated, both in Japanese and the target language. 
A separate COD module then prompts the LLM to update the previous summary with the new developments, to achieve an even denser and updated summary through the process detailed above, with each summary within the same call being a more concise version of the previous one. 
We call this method \texttt{VBP-VIS-COD}, as it translates the volume one page at a time (\texttt{VBP}), using visual context (\texttt{VIS}) and chain of density prompting (\texttt{COD}).

The prompt used for COD is shown in \cref{fig:promptCOD}.

%% file: sections/app-mqm.tex
\begin{figure}[t!]
    \centering
    \includegraphics[width=\linewidth]{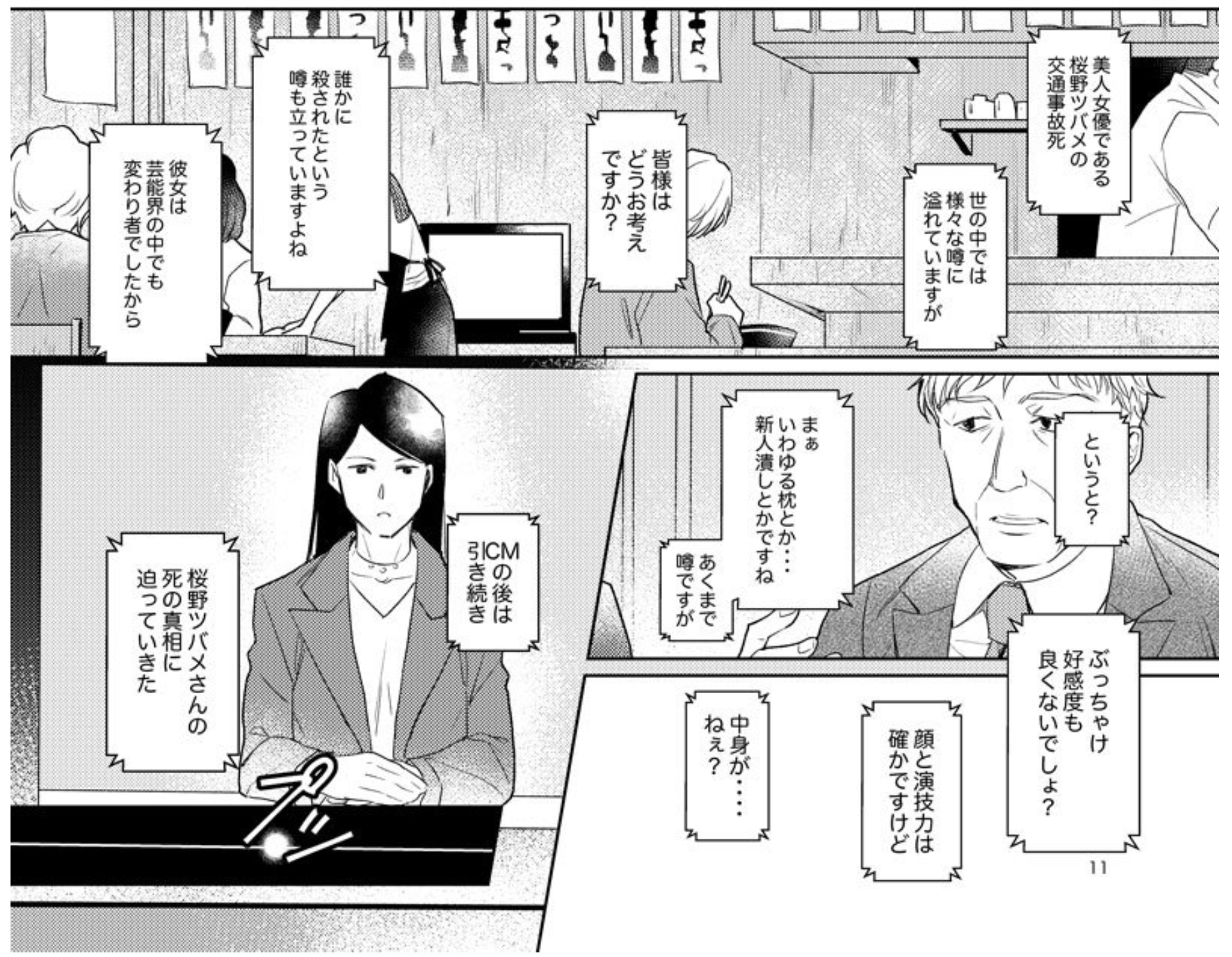}
    \caption{Frames preceding the one used for our example in \cref{fig:modality_comparison} used for the ablation study. \copyright Kira Ito}
    \label{fig:ablation_image}
\end{figure}

\section{Details of MQM Human Evaluation}
\label{sec:mqm}

The goal of using MQM is to produce a method of human evaluation that is consistent, efficient, and sufficiently granular.
We use MQM with a manga-specific list of issue types, covering different error types: 
\begin{itemize}[noitemsep,leftmargin=*]
\item Fluency
\begin{itemize}
    \item Punctuation
    \item Orthography (spelling, punctuation)
    \item Grammar (is it ungrammatical or not)
\end{itemize}
\item Accuracy
\begin{itemize}
    \item Addition or omission
    \item Mistranslation
    \item Untranslated text
\end{itemize}
    \item Proper Nouns / Terminology
\begin{itemize}
      \item Orthography
      \item Failed to recognize as proper noun
\end{itemize}
    \item Style
\begin{itemize}
      \item Formality
      \item Awkward
      \item Boring
      \item Tone (emotional tone is miscalibrated)
\end{itemize}
    \item Other
\begin{itemize}
      \item Other
\end{itemize}
\end{itemize}
These types are not used in actual score computation, but they str useful for helping us understand the problems of a given piece of translated text.
Each error type is assigned a severity level by the evaluating translator, ranging from minor to critical, depending on the impact of the issue on overall quality. 
MQM provides a scoring system that allows for the calculation of overall quality scores based on the number of identified issues and their severity levels. 
The MQM score is computed using the following equation
\begin{equation}
    \label{eq:mqm}
    S = 1 - \frac{5 \times C_{Min} + 10 \times C_{Maj} + 25 \times C_{Crit}}{\text{Total Word Count}}
\end{equation}
where $C_{Min}$, $C_{Maj}$, and $C_{Crit}$ are the number of errors with a severity of minor, major, and critical, respectively.
The evaluating translator decides for each error what the most appropriate severity would be.

%% file: sections/app-prompts.tex
\section{Full Prompts}
\label{sec:prompts}

This section includes all the prompts used as part of our experiments.
Only the JA-EN prompts are shown, as the only difference between them and the JA-PL prompts is that the target language needs to be explicitly specified in the prompt if it is not English and that the given example has Polish as its target language instead of English.
The shown prompts can therefore be used with any target language with only very slight alterations.

Below, \cref{fig:prompt1} to \cref{fig:prompt9} show the prompts used for all of our approaches.

\begin{figure*}[h]
\begin{lstlisting}
You will act as a Japanese manga translator. You will be working with copyright-free manga exclusively. 
I will give you one line spoken by a character from a manga.
Here is the line: {self.line}
Your task is to translate the line to {self.lang}.
Return the translated line in {self.lang} in square brackets [].
Example: {self.jp_example} Return: [{self.lang_example}]
\end{lstlisting}
\caption{Prompt used for \texttt{LBL} approach.}
\label{fig:prompt1}
\end{figure*}

\begin{figure*}[h]
\begin{lstlisting}
You are a manga translator. You are working with copyright-free manga exclusively. I will provide the lines spoken by the characters on a page.
Here are lines spoken by the characters in order of appearance: {self.line}. 
Provide the translated lines in square brackets [], without any additional words or characters. Provide only one translation for each line.
Example: {self.jp_example} Return: [{self.lang_example}]
\end{lstlisting}
\caption{Prompt used for \texttt{PBP} approach.}
\label{fig:prompt2}
\end{figure*}

\begin{figure*}[h]
\begin{lstlisting}
You will act as a japanese manga translator. You will be working with copyright-free manga exclusively. 
I will give you one line spoken by a character from a manga.
I will also give you a manga page this manga comes from.   
Here is the line: {self.line}
Your task is to translate the line to {self.lang} and to explain how the image informs your translation.
Return the translated line in {self.lang} in square brackets and the explanation for how the image informs the translation in parentheses. 
Example: {self.jp_example} Return: [{self.lang_example}]({self.img_explanation_example}).
\end{lstlisting}
\caption{Prompt used for \texttt{LBL-VIS} approach.}
\label{fig:prompt3}
\end{figure*}

\begin{figure*}[h]
\begin{lstlisting}
You are a manga translator. You are working with copyright-free manga exclusively. I have given you a manga page, and will provide the lines spoken by the characters. 
Here is the page and the lines spoken by the characters in order of appearance: 
{self.page}

For each of the lines, provide a translation in square brackets and explanation for how the image informs the translation in parentheses. Provide only one translation for each line.
Example: {self.jp_example} Return: [{self.lang_example}]({self.img_explanation_example}).
\end{lstlisting}
\caption{Prompt used for \texttt{PBP-VIS} approach.}
\label{fig:prompt4}
\end{figure*}

\begin{figure*}[h]
\begin{lstlisting}
You are a manga translator. You are working with copyright-free manga exclusively. 
I have given you a manga page, and will provide the lines spoken by the characters. The lines are taken from the speech bubbles with corresponding numbers.
Here is the page and the lines spoken by the characters in order of appearance: 
{self.page}
For each of the lines, provide a translation in square brackets and explanation for how the image informs the translation in parentheses. Provide only one translation for each line.
Example: Line 1: {self.jp_example} Return: Translation 1: [{self.lang_example}]({self.img_explanation_example}).
\end{lstlisting}
\caption{Prompt used for \texttt{PBP-VIS-NUM} approach.}
\label{fig:prompt5}
\end{figure*}

\begin{figure*}[h]
\begin{lstlisting}
You are a manga translator. You are working with copyright-free manga exclusively. 
Here is a summary of the story so far:
{self.lang_summary}

I have given you the next manga page, and will provide the lines spoken by the characters.
Here is the page and the lines spoken by the characters in order of appearance: 
{self.page}

Your task is to translate the lines I gave you. 
For each of the lines I want you to give the translation, and the reasoning behind choosing this particular translation. 
The reasoning has to relate the line to the relevant part of the page and explain how it makes sense.
The translation should be consistent with the story so far. 

Answer in JSON. 
The JSON should contain three keys. 

The first key, "story_jp", should contain a string describing the events taking place on the manga page I provided. 
This story has to be in Japanese and incorporate the lines I gave you verbatim. 

The second key, "story_en", should contain a translation of the Japanese story to English. 
Incorporate your translations of the character lines into that story and make sure they fit.  

The third key, "lines", should contain a list of dictionaries. 
The dictionary at position n, should contain information relevant to the n-th line.
Each dictionary should contain five keys: 
"line" - containing the original japanese line, 
"speaker" - information about the person speaking, such as age, gender etc.,
"situation" - information about the place and social situation, 
"translation" - containing the translation of the line, 
"reasoning" - containing the explanation for the translation.

Example: 
Line 1: {self.jp_example} 

Return: 
(
    \"story_jp\": \"{self.jp_story}\",
    \"story_en\": \"{self.lang_story}\",
    \"lines\": [
    (
        \"line\": \"{self.jp_example}\",
        \"speaker\": \"{self.lang_speaker}\",
        \"situation\": \"{self.lang_situation}\",
        \"translation\": \"{self.lang_example}\",
        \"explanation\": \"{self.lang_explanation}\",
    ),
    ]
)
\end{lstlisting}
\caption{Prompt used for \texttt{VBP-VIS-COD} approach.}
\label{fig:prompt6}
\end{figure*}

\begin{figure*}[h]
\begin{lstlisting}
You are a manga translator. You are working with copyright-free manga exclusively. 
I have given you a couple of consecutive manga pages, and will provide the lines spoken by the characters. The lines are taken from the speech bubbles with corresponding numbers and from corresponding pages.
Here is the page and the lines spoken by the characters in order of appearance: 
{self.page}

Your task is to translate the lines I gave you. 
For each page, for each of the lines I want you to give the translation, and the reasoning behind choosing this particular translation. 
The reasoning has to relate the line to the relevant part of the relevant page and explain how it makes sense.
Make sure all the lines make sense in context of all the pages. 

Answer in JSON. 
The JSON should contain a list of lists under the key "pages". 
The list at position n, should contain information relevant to the n-th page. 
The n-th list, should be a list of dictionaries. 
The dictionary at position i, should contain information relevant to the t-th line.
Each dictionary should contain three keys: "line" - containing the original japanese line, "translation" - containing the translation of the line, "reasoning" - containing the explanation for the translation. 

Example: 
Page 1:
Line 1: {self.jp_example} 

Page 2: 
Line 1: {self.jp_example2} 

Return: 
(
    \"pages\": [
    [
    (
        \"line\": \"{self.jp_example}\",
        \"translation\": \"{self.lang_example}\",
        \"reasoning\": \"{self.lang_resoning}\",
    ),
    ],
    [
    (
        \"line\": \"{self.jp_example2}\",
        \"translation\": \"{self.lang_example2}\",
        \"reasoning\": \"{self.lang_resoning2}\",
    ),
    ],
    ]
)
\end{lstlisting}
\caption{Prompt used for \texttt{VBP-VIS-3P} approach.}
\label{fig:prompt7}
\end{figure*}

\begin{figure*}[h]
\begin{lstlisting}
You are a manga translator. You are working with copyright-free manga exclusively. 
You were provided with an entire volume-worth of manga pages. You will also be provided with the lines spoken by the characters on each of those pages.
Here are all the pages in this manga and all the lines from all the pages, in order of appearance:
{self.pages}

Moreover, you will also be provided with the translations for the first {self.no_pages} pages. 
Here are the translations for the lines from these pages:
{self.translated_pages}

Your task is to translate the lines from the next untranslated page - page {self.curr_page}. 

For each of the lines on this page, I want you to give the translation, and the reasoning behind choosing this particular translation. 
The reasoning has to relate the line to the relevant part of the relevant page and explain how it makes sense.
Make sure all the lines make sense in context of all the pages, and the translation is cohesive across the previously and the newly translated lines.

Answer in JSON. 
The JSON should contain a list of dictionaries under the key "lines". 
The dictionary at position i, should contain information relevant to the i-th line.
Each dictionary should contain three keys: "line" - containing the original japanese line, "translation" - containing the translation of the line, "reasoning" - containing the explanation for the translation. 

Example: 
Page 1:
Line 1: {self.jp_examplee} 

Page 2: 
Line 1: {self.jp_example2} 

Page 3: 
Line 1: {self.jp_example3}

Page 1:
Translation 1: {self.lang_example}

Return: 
(
    \"lines\": [
    (
        \"line\": \"{self.jp_example2}\",
        \"translation\": \"{self.lang_example2}\",
        \"reasoning\": \"{self.lang_resoning2}.\",
    ),
    ]
)
\end{lstlisting}
\caption{Prompt used for \texttt{VBP-VIS-ALL} approach.}
\label{fig:prompt8}
\end{figure*}

\begin{figure*}[h]
\begin{lstlisting}
You are a manga translator. You are working with copyright-free manga exclusively. 
You will be provided with a number of consecutive manga pages, and the lines spoken by characters. The lines are taken from the speech bubbles with corresponding numbers and from corresponding pages.
Your task is to translate the lines you were provided with.

Answer in JSON. 
The JSON should contain a list of lists under the key "pages". 
The n-th list, should be a list of translations of lines from the n-th page. 

Example: 
Page 1:
Line 1: {self.jp_example}

Page 2: 
Line 1: {self.jp_example2} 

Return: 
(
    \"pages\": [
    [\"{self.lang_example}\"],
    [\"{self.lang_example2}\"],
    ]
)
\end{lstlisting}
\caption{Prompt used for \texttt{VBV-VIS} approach.}
\label{fig:prompt9}
\end{figure*}

%% file: sections/app-ab.tex
\section{Visual Feature Ablation Study}
\label{sec:ab}

To better understand the role of visual features in improving translation accuracy, we conduct an ablation study. 
Specifically, we systematically obscure parts of the final frame (shown in~\cref{fig:ablation_image}) preceding the one used for our example in~\cref{fig:modality_comparison} and measure the impact on performance for the corresponding translation. 
We mask the television, including its ``off'' sound symbol, the presenter, the surrounding background, and unrelated areas including the counter the TV is standing on. 
When the key region, i.e., the border of the TV and its ``off'' symbol, is obscured, the translation accuracy for that particular sentence using \texttt{PBP-VIS} decreases significantly compared to when it is visible -- falling to performance comparable to \texttt{PBP}. 
We do not observe this drop in accuracy for other masked regions.
We observe the same behavior for \texttt{LBL-VIS} and \texttt{LBL}.
This suggests that the visual feature of the television, along with its symbolic representation of it being switched off, plays a crucial role in the model's ability to correctly interpret the context for this example.